\def\year{2022}\relax
\documentclass[letterpaper]{article} % DO NOT CHANGE THIS
\usepackage{aaai22}  % DO NOT CHANGE THIS
\usepackage{times}  % DO NOT CHANGE THIS
\usepackage{helvet}  % DO NOT CHANGE THIS
\usepackage{courier}  % DO NOT CHANGE THIS
\usepackage[hyphens]{url}  % DO NOT CHANGE THIS
\usepackage{graphicx} % DO NOT CHANGE THIS
\usepackage{natbib}  % DO NOT CHANGE THIS AND DO NOT ADD ANY OPTIONS TO IT
\usepackage{caption} % DO NOT CHANGE THIS AND DO NOT ADD ANY OPTIONS TO IT
\DeclareCaptionStyle{ruled}{labelfont=normalfont,labelsep=colon,strut=off} % DO NOT CHANGE THIS
\usepackage{graphicx}
\usepackage{amsmath}
\usepackage{amsthm}
\usepackage{amssymb}
\usepackage{booktabs}
\usepackage{algorithm}
\usepackage{array}
\usepackage{subfigure}
\usepackage{multirow}
\usepackage{threeparttable}
\usepackage{tabularx}
\usepackage{booktabs}
\usepackage{colortbl}
\usepackage{xcolor}
\usepackage{stfloats}
\usepackage[noend]{algpseudocode}
\usepackage{mathtools}
\usepackage{makecell}
\usepackage{setspace}

\usepackage[switch]{lineno}
\usepackage{newfloat}
\usepackage{listings}
\floatstyle{ruled}
\newfloat{listing}{tb}{lst}{}
\floatname{listing}{Listing}
\title{KNN-BERT: Fine-Tuning Pre-Trained Models with KNN Classifier}
\author{
    Linyang Li \equalcontrib,
    Demin Song \equalcontrib,
    Ruotian Ma,
    Xipeng Qiu \thanks{corresponding author},
    Xuanjing Huang
}
\affiliations{
    School of Computer Science, Fudan University\\
    Shanghai Key Laboratory of Intelligent Information Processing, Fudan University\\
    
    \texttt{\{linyangli19,dmsong20,xpqiu\}@fudan.edu.cn}
}

\usepackage{bibentry}
\begin{document}

\maketitle

\begin{abstract}
Pre-trained models are widely used in fine-tuning downstream tasks with linear classifiers optimized by the cross entropy loss, which might face robustness and stability problems.
These problems can be improved by learning representations that focus on similarities in the same class and contradictions in different classes when making predictions.
In this paper, we utilize the K-Nearest Neighbors Classifier in pre-trained model fine-tuning.
For this KNN classifier, we introduce a supervised momentum contrastive learning framework to learn the clustered representations of the supervised downstream tasks.
Extensive experiments on text classification tasks and robustness tests show that by incorporating KNNs with the traditional fine-tuning process, we can obtain significant improvements on the clean accuracy in both rich-source and few-shot settings and can improve the robustness against adversarial attacks.
\footnote{all codes is available at https://github.com/LinyangLee/KNN-BERT}
\end{abstract}

\section{Introduction}

Pre-trained language models exemplified by BERT \cite{bert} have been widely applied in fine-tuning downstream text classification tasks.
It is commonly used to fine-tune the pre-trained model with the cross entropy loss \cite{rumelhart1986learning} that calculates the KL-divergence between the one-hot vectors of labels and the model output predictions and then make predictions using linear classifiers \cite{radford2019language,bert,liu2019roberta,yang2019xlnet,lan2019albert}.

Still, such a standard process has its shortcomings:
\textbf{(A)} the cross entropy loss may lead to poor generalization performance as pointed out by \citet{liu2016large,cao2019learning} and may lack robustness against noisy labels \cite{zhang2018generalized,sukhbaatar2014training} and adversarial samples \cite{goodfellow2014explaining,nar2019cross}. Also, in fine-tuning BERT, the cross entropy loss may be unstable especially with limited data \cite{dodge2020fine,zhang2020revisiting}.  
\textbf{(B) }
On the other hand, making predictions through linear classifiers added directly on top of the pre-trained models may face the overfitting problem especially when the training data is limited \cite{snell2017prototypical,zhang2020revisiting}.

To tackle the above shortcomings, it is intuitive to build better representations in pre-trained language models and make predictions based on classifiers that have better generalization abilities.

% use KNN
Therefore, in this paper, instead of simply using a linear classifier to do the prediction, we utilize the classic K-Nearest Neighbors classifier to make predictions based on the training sample representations.
The classic KNN classifier that makes predictions based on counting the top-K similar samples has been neglected for a long time since end to end neural networks have achieved great success in the computer vision field \cite{he2016deep} as well as the natural language processing field \cite{vaswani2017attention,bert}.
However, when the representations have been well-learned through the massive calculation of the masked language model task pre-training, 
it is intuitive to revisit and utilize the K-Nearest Neighbor classifier that makes predictions based on the representation similarity.
The KNN classifier makes predictions based on the similarity between representations.
Therefore, the decision boundary is tighter within the same class, therefore, altering the representation to an incorrect class is more difficult which can improve model robustness.
On the other hand, the KNN classifier make predictions based on the anchors of the multiple training samples which are well-learned representations from the BERT model.
Therefore, utilizing KNN classifier can make better use of the semantic representations of the pre-trained models than simply using linear classifiers to draw decision boundaries.
% 这个 问题 emmm

To train the representations that are clustered within the same class for the KNN classifier, it is intuitive to use contrastive learning based training strategies.
The goal is to construct a tight cluster of the representations within the same class while keep the clusters of different classes at distance.
With the label information from the downstream task dataset, we introduce a class-wise supervised contrastive learning framework to cluster the representations.
Based on traditional constrastive learning framework, we use the class-wise positives drawn from the same class of the given example instead of using limited augmentation-based methods to construct positives.
These class-wise positives are relatively more abundant and useful compared with augmentation-based positives and they can also be diversified in semantics.

To make use of the class-wise positives, we incorporate the momentum contrast learning framework (MoCo) \cite{he2020momentum} and introduce a sampling strategy that collect both similar positives and hard to distinguish positives to learn representations that can be tightly clustered within the same class while can be distant between different classes.
The momentum contrast framework introduces a queue-based optimization process to update the representations of the negatives which makes it possible to make use of massive negatives.
In our usage of contrastive learning, incorporating the queue-based momentum contrast allows the usage of massive positives and negatives which is intuitive in using class-wise positives. 
That is, we put a large number of negatives in the queue and update the representations of these negatives via the momentum contrast.
Meanwhile, we update the representations of multiple positives via gradient decent.

For the representation learning of the positives, we are hoping that (1) the cluster of samples is tight within the same class; (2) the clusters are distant between classes.
Therefore, when updating the representations of the class-wise positives, we introduce a sampling strategy that consider the most similar and least similar positives to get better cluster representations.
Updating the most similar positives can draw near the representations within the same class while updating the least similar positives will push away representations from different classes.

We construct extensive experiments to test the generalization and robustness ability of our contrastive-learned representations for the KNN classifier.
We test rich-source and low-source text classification tasks on the GLUE benchmark;
we then test the robustness of the KNN classifier by using the classifier to defend against strong substitution-based adversarial attack methods such as Textfooler \cite{jin2019textfooler} and Bert-Attack \cite{li2020bert}. 
Experiment results indicate that the KNN classifier can (1) improve the performances by a considerable margin in text classification tasks; (2) improve the defense ability against adversarial attacks significantly. 

To summarize our contributions:

\begin{itemize}
    \item We introduce the idea of utilizing traditional KNN classifier in downstream task fine-tuning of pre-trained models and use contrastive-learning to learn the representations for the KNN classifier.
    
    \item We make use of class-wise positives and negatives and introduce a sampling strategy that consider most and least similar positives for the contrastive learning process.
    
    \item We incorporate a momentum contrast based framework to allow multiple positives and negatives in the contrastive learning process.

    \item Extensive experiments show the effectiveness of the proposed contrastive learning framework for the KNN classifier in both model generalization ability and model robustness.
    
\end{itemize}

\section{Related Work}

\subsection{Utilizing the KNN Classifier in PTMs}

The K nearest neighbor classifier is a traditional algorithm that makes predictions based on representation similarities.
While pre-trained models (PTMs) \cite{bert,radford2018improving,lan2019albert,liu2019roberta} have been widely applied, the idea of using nearest neighbors in pre-trained models is also explored.
\citet{khandelwal2019generalization} uses nearest neighbors to augment the language model predictions by using neighbors of the predictions as targets for language model learning.
\citet{kassner2020bert} applies nearest neighbors as additional predictions to boost the question answering task.
These methods use nearest neighbors to find augment samples based on the pre-trained language models rather than using the KNN classifier as the decision maker.

On the other hand, making predictions based on the nearest neighbors can be used in improving model robustness. 
\citet{papernot2018deep} explores the possibility of using nearest neighbors to make decisions instead of using linear classifiers in the computer vision field, showing that classification results based on near neighbors are more resilient to adversarial attacks \cite{goodfellow2014explaining,CarliniW16a}.

\subsection{Contrastive Learning}

Contrastive learning \cite{hadsell2006dimensionality,chen2020simple} is a similarity-based training strategy that has been widely used \cite{hjelm2018learning,sermanet2018time,tschannen2019mutual}.
The formulation of the contrastive loss is mainly based on the noise contrastive estimation loss \cite{gutmann2010noise,mnih2013learning} or the N-pair losses \cite{sohn2016improved}, which is also closely related to the metric distance learning and triplet losses \cite{schroff2015facenet,weinberger2009distance}.

Recent contrastive learning framework is widely used in self-supervised tasks \cite{he2020momentum,chen2020simple}, while the contrastive losses can also be used in a supervised scenario with minor modification to the loss function \cite{khosla2020supervised,gunel2020supervised}.
These supervised contrastive learning losses are added as an additional task in the normal training, the inference process is still based on linear classifiers.

On the other hand, the contrastive losses require a large number of negative samples.
\citet{he2020momentum} introduces a momentum contrast approach that incorporates a queue of data samples and the encoder of these samples is updated in a momentum process.
The end-to-end update process uses the samples in the current mini-batch therefore limited by the GPU memory size; while the memory bank approach \cite{wu2018unsupervised} store the representations that cannot be carefully updated.
The momentum contrast approach makes it possible to train the supervised contrastive loss of multiple positives and negatives in large-scale pre-trained language model fine-tuning.

\section{KNN-BERT}
We propose KNN-BERT that utilizes the KNN classifier when using pre-trained models exemplified by BERT as the representation encoder.
We illustrate the KNN-BERT by describing (1) the KNN classifier usage; (2) the training process of the representations for the KNN classifier.

\begin{figure*}[ht]
\centering
\includegraphics[width=1.0\linewidth]{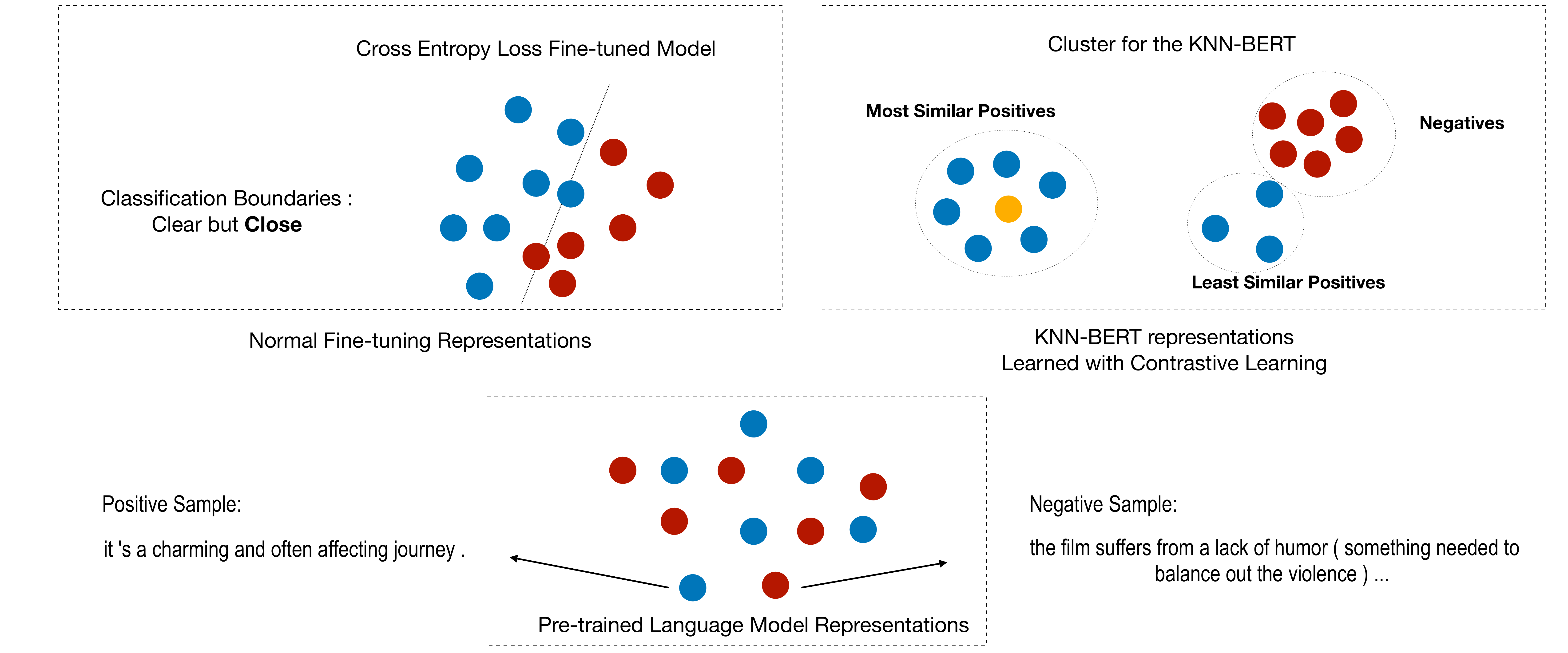}
\centering
\caption{An illustration of using contrastive learning methods to build clusters for downstream classification tasks. We use dots of different colors to denote different classes.
The most similar positives are the major cluster while the least similar positives are the data points that are closer to the negative classes.
The KNN classifier use an anchor-based prediction strategy unlike previous linear classifiers, we use contrastive loss to make the clusters tighter and draw the least similar samples towards the major cluster.}
\label{fig:example}
\end{figure*}

\subsection{KNN Classifier}

We combine the normal linear classifier with the KNN classifier and use the weight-averaged logits as the final prediction logits.
Suppose the encoded representation is $q$ with label $Y_q$ and the linear classifier is $F(\cdot)$;
we use $k_i$ with label $Y_{k_i}$ to denote the $K$ nearest neighbors measured by the cosine similarity.

The KNN logits is a voted result denoted as $KNN(q)$.

With weight ratio $\phi$, 
the final prediction score $s$ is calculated by: 

\begin{align}
    s = (1-\phi) \text{Softmax}(F(q)) + \phi \text{KNN}(q)
\end{align}

Here, the linear classifier $F(\cdot)$ is learned by traditional cross entropy loss.
For the kNN classifier learning, we illustrate our proposed contrastive learning framework in the following section.

% Use \bibliography{yourbibfile} instead or the References section will not appear in your paper

% \newpage

\subsection{Contrastive Learning for KNN}

In order to train representations for the KNN classifier in fine-tuning pre-trained models, we introduce a supervised contrastive learning framework that makes use of label information to construct positive and negative samples.

Derived from the info-nce loss, we consider a supervised contrastive loss function $\mathcal{L}_\textbf{sc}$:

\begin{align}
\begin{split}
          \mathcal{L}_\textbf{sc} = \frac{1}{M} \sum_{k_j \in k_+} \big( - \text{log} \frac{ \text{exp}(q \cdot k_{j} / \tau )}{\sum\limits_{k_i \in \{ k_{-}, k_{j} \} }    \text{exp}(q \cdot k_{i} / \tau)    } \big)
          \label{eq:scloss}
\end{split}
\end{align}

Here, $k_+$ is the set of $M$ samples that have the label with the given query sample $q$ and $k_{-}$ is the set of samples from different classes.
Such a loss function could narrow down the gap between the query and the positive samples and push away the query and the negatives.
Considering that the positive samples could be diversified since they are from the same class but the representations possess various semantic information encoded by pre-trained models,
it is important to determine which positives should be used in calculating the similarities with the given query, otherwise, the learned representations may not be tightly clustered.

Therefore, we aim to learn the clusters by (1) tighten the cluster of samples of the same class; (2) push away samples from different classes.

As seen in Figure \ref{fig:example}, we calculate similarities between the most similar positives and the query to build a tighter cluster by narrowing the gap of these most similar samples with the query.
On the other hand, we select the least similar positives and draw them towards the query sample.
Optimizing the gap between the least similar positives and the query sample is similar with using hard-negatives for better clustering, so we can also name these positives as hard-positives.

Therefore, we select $M_{m}$ most similar positives $k_m$ and $M_{l}$ least similar positives $k_l$ from positives set $k_+$ and only update these selected positive sample representations. 
Calculating all positives might sabotage the semantic information which may not be related to the classification representations and hurt the classification results since the class-wise positives can be significantly different with the query sample.
The proportion of the selected most and least similar positives would play a vital role in the cluster learning process, which will be discussed in the later section.

Since we only update these selected positives, we can re-write the contrastive loss function to:

\begin{align}
\begin{split}
          \mathcal{L}^{'}_\textbf{sc} = \frac{1}{M_m+M_l}  \sum_{k_j \in \{ k_{m},k_{l} \} }  \big(      - \text{log}  \frac{ \text{exp}(q \cdot k_{j} / \tau )}{\sum\limits_{k_i \in \{ k_{-}, k_{j} \} }    \text{exp}(q \cdot k_{i} / \tau)    } \big)
          \label{eq:finalloss}
\end{split}
\end{align}

\subsection{Connections with Traditional Contrastive Learning}

Contrastive learning \cite{hadsell2006dimensionality,chen2020simple} is to train a representation (denoted as $q$) that has positive keys (denoted as $k_{+}$) and negative keys (denoted as $k_{-}$).
When the similarity is measured by the dot-product, a \textbf{c}ontrastive loss with one positive key with multiple negative keys ($N$ negatives) is considered as:
\begin{align}
      \mathcal{L}_\textbf{c} = - \text{log} \frac{ \text{exp}(q \cdot k_{+} / \tau )}{\sum\limits_{k_i \in \{ k_{-}, k_{+} \} }    \text{exp}(q \cdot k_{i} / \tau)    } 
\label{eq:closs}
\end{align}
Here $\tau$ is the temperature hyper-parameter, and $\{k_{-},k_+ \} $ is the sum of over one positive $k_+$ and $N$ negatives $k_i \in k_{-}$, which is $N+1$ samples in total. This form of loss is closely related to the widely used information noise contrastive estimation \cite{oord2018representation}.
This form of loss is widely used in self-supervised contrastive learning tasks \cite{sohn2016improved,oord2018representation,henaff2020data,baevski2020wav2vec} where the positive can be constructed using data augmentation methods \cite{chen2020simple}.

Compared with the traditional contrastive learning method, tasks such as text classification are supervised tasks where supervised contrastive learning is explored in the language understanding tasks \cite{gunel2020supervised}.
The major difference is that supervised contrastive learning allows multiple usage of positives since the positives can be drawn from the same class with the query sample.

Based on the self-supervised and supervised contrastive learning frameworks, we build our proposed contrastive learning framework for the KNN-BERT model.

\subsection{Optimizing with Momentum Contrast}

As illustrated in Eq. \ref{eq:finalloss}, we are using multiple positives and a large number of negatives in calculating the contrastive loss,
therefore, we utilize a momentum contrast framework to update the positives and negatives for better representation clustering.

In the contrastive learning training process, incorporating massive negatives can help better sample the underlying continuous high-dimensional space of the encoded representations.
Therefore, the momentum contrast framework (MoCo) is introduced \cite{he2020momentum} to consider very large amount of negatives using a queue-based update strategy.

In the momentum contrast framework, there are two separate encoders: query encoder and key encoder. 
The query encoder is updated by using the gradient descent of the query samples.
The optimization of the key encoder is solved by a momentum process using the parameters from the query encoder as illustrated below:

\begin{align}
    {\theta}_k \gets m {\theta}_k + (1-m) {\theta}_q
\end{align}

Here ${\theta}_q$ and ${\theta}_k$ are the encoders while only the query encoder ${\theta}_q$ is updated by gradients through back-propagation.

The negative representations are first pushed into the recurrent queue and only the samples in the end of the queue is updated by encoding with the key encoder after the key encoder is updated by the momentum process based on the query encoder.
Through the momentum update process, the constrastive learning process can consider a great amount of positives and negatives since the process does not need to calculate the gradients on all positives and negatives.

Different from the traditional Moco framework where the positive sample is updated based on gradients, we have large amounts of both positives and negatives in the supervised contrastive learning setting.
We simply push all these samples in to the queue and construct the positives and negatives based on the label of the query sample.

% Under the supervised contrastive learning setting, we can update multiple positives using a large number of negatives based on the MoCo framework, which can construct tighter clusters for the KNN classification process.

\subsection{Combined Training}

In the pre-trained model fine-tuning exemplified by text classification tasks, the representations are the \texttt{[CLS]} tokens used for text classification tasks.
We use the $l_2$ normalization over these representations since normalization methods are widely used in contrastive learning methods and have been proved useful through empirical results. 
Therefore, the queries and their corresponding positives and negatives are the representations of the BERT encoder output \texttt{[CLS]} tokens.

We add the contrastive loss along with the original cross entropy loss for $C$ classes in the fine-tuning process to make used of the label information in a more direct way.
\begin{align}
    \mathcal{L}_{ce} = - \sum_{c=1}^C y_c  \cdot log (F(q)_c)
\end{align}
Here, $F(\cdot)$ is the linear classifier and $F(q)_c$ is the probability of the $c_{th}$ class correspondingly.

Therefore, the final training loss is:
\begin{align}
    \mathcal{L} = (1-\lambda) \mathcal{L}_{ce}  + \lambda \mathcal{L}^{'}_{\textbf{sc}}
\end{align}

\begin{table*}[ht]
\setlength{\tabcolsep}{10pt}
\centering
\small
\begin{tabular}{lcccccccc}
\toprule
Methods & RTE & MRPC  & QNLI & MNLI & SST-2 & IMDB & AG's News  \\
& \\
\midrule
BERT  &  65.34 & 88.99 & 91.37 & 84.51 &  92.72 & 93.50 &  90.13 \\
SCL & 67.87 & 87.97 & 90.99 & 84.35 & 92.43 & 92.65& 89.82 \\
SCL-MoCo & 71.11 & 88.90 & 91.26 & 84.50 & 92.70 & 93.50& 90.11 \\
KNN-BERT & \textbf{72.56} & \textbf{91.22} &  \textbf{91.74} & \textbf{84.69} & \textbf{93.11} & \textbf{93.62} &  \textbf{90.56} \\
\bottomrule
\end{tabular}
\caption{Main Results on full-data text classification tasks and sentence pair classification tasks.}
\label{tab:main-results}
\end{table*}

% \newpage

\section{Experiments}

\subsection{Datasets}

We use several text classification datasets to evaluate the effectiveness and robustness of our proposed KNN-based classifier.

We use several datasets in the GLUE benchmark \cite{wang2018glue}: the Recognizing Textual Entailment dataset (RTE \cite{rte}; Microsoft Research Paraphrase Corpus dataset (MRPC \cite{dolan-brockett-2005-automatically}); Question Natural Language Inference dataset (QNLI \cite{rajpurkar-etal-2016-squad});  Multi-genre Natural Language Inference dataset (MNLI \cite{mnli}) and the Standard Sentiment Treebank dataset (SST-2 \cite{socher-etal-2013-recursive}).

In testing the text classification models, we have two experiment settings: we train the model with the full training dataset and test on the validation set; we also set a few-shot setting with only a small portion of the training set. 
We sample a test set and a development set from the given development set following \cite{gunel2020supervised}.

We also use the IMDB movie review dataset \cite{maas2011learning} and the AG's News news-genre classification dataset \cite{zhang2015character} to test the generalization ability as well as the model robustness.
We use the well-known substitution-based adversarial attack methods, Textfooler \cite{jin2019textfooler} and Bert-Attack \cite{li2020bert} to attack our KNN classifier.

\subsection{Implementations}

We run the experiments based on the BERT-BASE \cite{bert} and RoBERTa Large model \cite{liu2019roberta} using the Huggingface Transformers.
We use the standard fine-tuning hyper parameters with learning rate set to 2e-5 and batch-size set to 8 and optimize using the Adam optimizer. 
The parameters are not particularly tuned, we only use the parameters provided by the Transformers toolkit \footnote{https://github.com/huggingface/transformers}.
In the experiments that concern the contrastive learning process, we search for proper hyperparameters.
The size of the queue is 32000, while in the tasks with a small size of training set we put the entire dataset into the queue.
We set the momentum update parameter $m=0.999$ with the temperature $\tau = 0.07$.
We set the positives number $M_t$ and $M_n$ considering the training set size of different tasks which are discussed later. 
The ratio between the contrastive and the linear classifier in both training process and inference time is tuned through a sweep for $\lambda$ and  $\phi$, which we further discuss in the later sections. 

For the robustness experiments, we use Textfooler \cite{jin2019textfooler} and BERT-Attack \cite{li2020bert} as the adversarial attack methods to attack the downstream task classifiers.
We use the TextAttack Toolkit \cite{morris2020textattack} to implement the attack methods and test the performances against adversarial attacks using our KNN-based classifier.

\subsection{Baselines}

We compare our KNN-based classifier with several contrastive learning methods.
We train these methods using the same parameters with our KNN-based approach for a fair comparison.

 \textbf{SCL}
We first construct a supervised contrastive loss involved training baseline which is similar to \citet{gunel2020supervised}.
The supervised contrastive loss is similar to Eq. \ref{eq:scloss} where the positive and negatives are randomly selected in the minibatch. 
The SCL method is contrained by the memory size of the GPU memory so the number of positives and negatives is limited.

\textbf{SCL-MoCo}
We then construct a more delicate baseline that incorporates the contrastive loss using the MoCo \cite{he2020momentum} framework. 
That is, the negatives are drawn from the queue which is significantly larger than the batch size.

\begin{table}
\setlength{\tabcolsep}{8pt}
\centering
\small
\begin{tabular}{lcccc}
\toprule
Methods & RTE & MRPC  & QNLI  & SST-2   \\
& \\
\midrule
BERT  &  66.4 & 88.9 & 90.5 & 93.5 \\
KNN-BERT & \textbf{70.2} & \textbf{89.1} & \textbf{90.8} &  \textbf{93.5} \\
\bottomrule
\end{tabular}
\caption{Main Results on the test server of GLUE benchmark using models checkpoints based on the best development set results.}
\label{tab:main-results-test}
\end{table}

\subsection{Main Results}

We propose a KNN-based classifier trained with MoCo-based contrastive learning framework and we test on the widely acknowledged GLUE benchmark as shown in Tab.\ref{tab:main-results}.
We observe that when using the KNN classifier, the model performances have an average improvement of 1.39 points compared with the BERT baseline.
We also test the KNN classifier on the test server of the GLUE benchmark \footnote{https://gluebenchmark.com/} as shown in Tab.\ref{tab:main-results-test}.

We compare our KNN-BERT method with several contrastive learning baselines.
As seen, when we use the contrastive learning loss in the training stage with negatives sampled from the minibatch, the performances improve by a small margin compared with the BERT baselines.
Further, when we only use the MoCo training as an additional loss in the training process, the model performances are still behind the KNN-BERT method that is trained with the multiple positives contrastive loss and makes predictions with both the linear classifier and the KNN classifier.
We can conclude that incorporating the KNN classifier in the language model fine-tuning is beneficial when the model is trained with a contrastive loss that cluster the data space for each class.

Further, we can observe that the tasks with relatively low-source training sets have more considerable improvements which indicate that the contrastive learned clusters can help improve limited data scenarios, which we will further discuss in the few-shot experiments. 

\subsection{Few-Shot GLUE Results}

As mentioned, we observe that the contrastive loss based KNN classifier can achieve better results in low-source tasks.
Therefore, we construct a few-shot experiment using limited data for the downstream tasks.

As seen in Tab.\ref{tab:main-fewshot-results}, both BERT and RoBERTa models can be improved by the KNN classifier when the training set has only 100 or 1000 training samples in the SST-2, QNLI and IMDB dataset.
The few-shot setting constrains the performances of language model fine-tuning compared with the rich-source fine-tuning,
while the KNN classifier can gain a more significant improvement in the few-shot settings compared with the rich-source fine-tuning.
Plus, we can observe that the KNN classifier have a relatively small variance, indicating that the performance is more stable.

We assume that when the training data is limited, the linear classifier would face a serious overfitting problem.
The similarity-based KNN classifier, on the other hand, considers more connections between the samples in the same class, which contributes to the improvements over the few-shot experiments.
Compared with the baseline supervised contrastive learning methods, using the KNN classifier to make predictions can achieve a higher performance.
We can assume that with limited data, the representations xxx 
%  PTMs help few-shot KNN classifier xxx

\begin{table}
\setlength{\tabcolsep}{6pt}
\centering
\scriptsize
\begin{tabular}{clccccccc}
\toprule
Num. & Methods  & SST-2 & QNLI & IMDB \\
\midrule

\multirow{8}{*}{100} &BERT   & 78.90(3.31) & 65.76(29.87) & 73.38(13.39) \\
& SCL  & 75.96(9.37) & 65.27(26.20) & 73.65(6.52) \\
& SCL-MoCo  & 79.63(5.85) & 68.14(0.24) & 74.82(13.98) \\
& KNN-BERT  & \textbf{81.36}(5.85) & \textbf{70.52}(0.45) & \textbf{79.56}(1.95) \\
\cline{2-5}

\rule{0pt}{2.5ex} &RoBERTa  & 92.16(0.83) & 70.40(47.72) & 92.66(0.20)  \\
& SCL  & 90.00(1.88) & 71.39(53.64) & 92.21(0.48) \\
& SCL-MoCo  & 91.14(1.01) & 72.90(57.81) & 92.71(0.57) \\
&KNN-RoBERTa & \textbf{93.20}(0.10) & \textbf{76.00}(37.26) & \textbf{93.68}(0.41) \\
\midrule

\multirow{8}{*}{1000} &BERT  & 88.30(0.63) & 76.26(1.25) & 88.82(0.08) \\
& SCL  & 89.40(0.06) & 77.16(0.52) & 88.53(0.07) \\
& SCL-MoCo  & 88.58(0.85) & 77.34(0.63) & 89.11(0.12) \\
&KNN-BERT   & \textbf{89.96}(0.37) & \textbf{77.68}(0.89) & \textbf{91.68}(0.53) \\

\cline{2-5}

\rule{0pt}{2.5ex} & RoBERTa   & 93.26(0.28) & 85.32(1.57) & 94.47(0.03)  \\
& SCL  & 93.49(0.23) & 87.15(3.71) & 94.24(0.02) \\
& SCL-MoCo  & 93.90(0.28) & 85.98(0.17) & 94.01(0.18) \\
&KNN-RoBERTa   & \textbf{94.04}(0.13) & \textbf{87.32}(0.09) & \textbf{96.08}(0.55) \\
\bottomrule
\end{tabular}
\caption{Few-Shot Results on the constructed test set.
We run 5 times using different seeds and use the averaged performance with variance given in the parentheses.}
\label{tab:main-fewshot-results}
\end{table}

\subsection{Model Robustness against Adversarial Attacks}

The robustness of neural networks has raised more and more concerns while these powerful models are widely applied.
To explore the robustness of our KNN classifier against strong adversarial attack methods, we construct a robustness experiment to put our KNN classifier as the target model for the strong attacking methods.

We use two different settings: (1) the predictions are made by both the linear classifier and the KNN classifier ($\phi = 0.5$)  (2) predictions are only made by the KNN classifier ($\phi=1$).

As seen in Tab.\ref{tab:robust-results},
utilizing the KNN classifier is helpful in obtaining a higher accuracy when attacked by strong adversarial attack methods.
Since the attacking process is an iterative searching process, it becomes harder to find proper substitutions as adversarial examples when the distance between classes is larger.
% 先说 KNN 的advantage 
The comparison between the KNN-only classifier($\phi=1.0$) and the KNN \& Linear combined classifier($\phi=0.5$) indicates that the linear classifier is not robust even when the model has been trained with contrastive losses, which reveals a strong advantage of utilizing the KNN classifier in the pre-trained model fine-tuning.
% 再说 defense 里面 的 作用 
Previous methods introduced in improving model robustness focus on finding the data space of the adversaries and defend via adversarial training based methods.
While in the KNN classifier, the robustness improvements are obtained by the closer distributions over the clean examples from the training set serving as anchors, which could provide strong defense results against adversarial examples.

\begin{table}
\centering
\small
\begin{tabular}{lcccccccc}
\toprule
Methods & Origin & Textfooler & BERT-Attack \\
\midrule
\multicolumn{2}{c}{\bfseries IMDB }\\
\midrule
BERT    & 93.7 & 24.7 & 17.3 \\
KNN($\phi=0.5$) & 94.3 &  44.7 & 25.7 \\
KNN($\phi=1.0$) & 94.3 & \textbf{52.7} & \textbf{30.0}  \\

\midrule
\multicolumn{2}{c}{\bfseries AG's News }\\
\midrule
BERT  & 88.0 & 22.0 & 21.7 \\
KNN($\phi=0.5$) & 90.6 & 22.7 & 37.3 \\
KNN($\phi=1.0$) & 90.3 & \textbf{47.7} & \textbf{42.7} \\
\bottomrule
\end{tabular}
\caption{Robustness experiments tested on strong adversarial attack methods (KNN is the the KNN-BERT method).
The metric is the after-attack accuracy.}
\label{tab:robust-results}
\end{table}

\subsection{Ablations}

We conduct an extensive ablation study to explore the effectiveness of our proposed contrastive learning framework.

\begin{figure*}
\centering
\subfigure[selection of $K$]{
\begin{minipage}[t]{0.5\linewidth}
\includegraphics[width=1.0\linewidth]{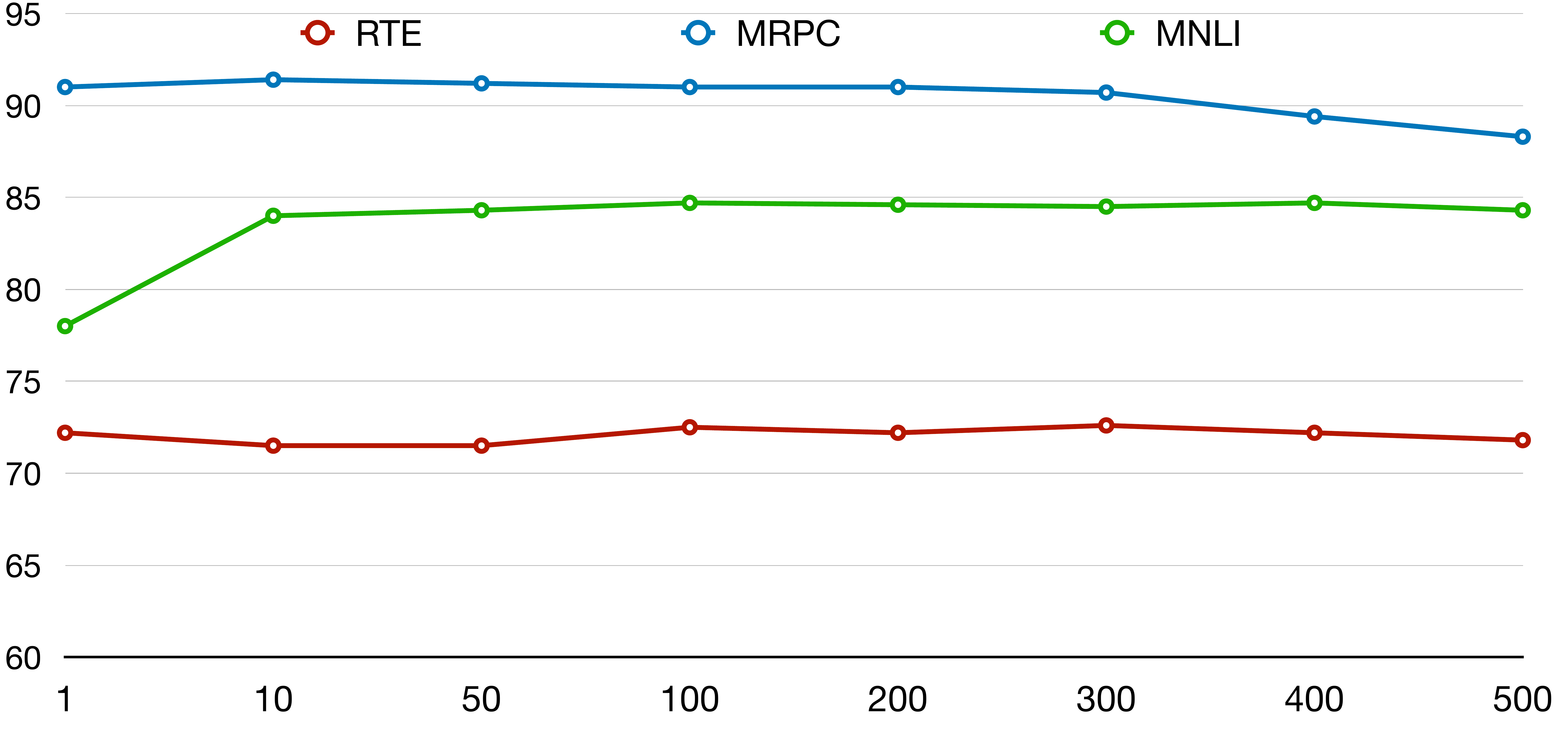}
%\caption{fig1}
\end{minipage}%
}%
\subfigure[selection of $\phi$]{
\begin{minipage}[t]{0.5\linewidth}
\includegraphics[width=1.0\linewidth]{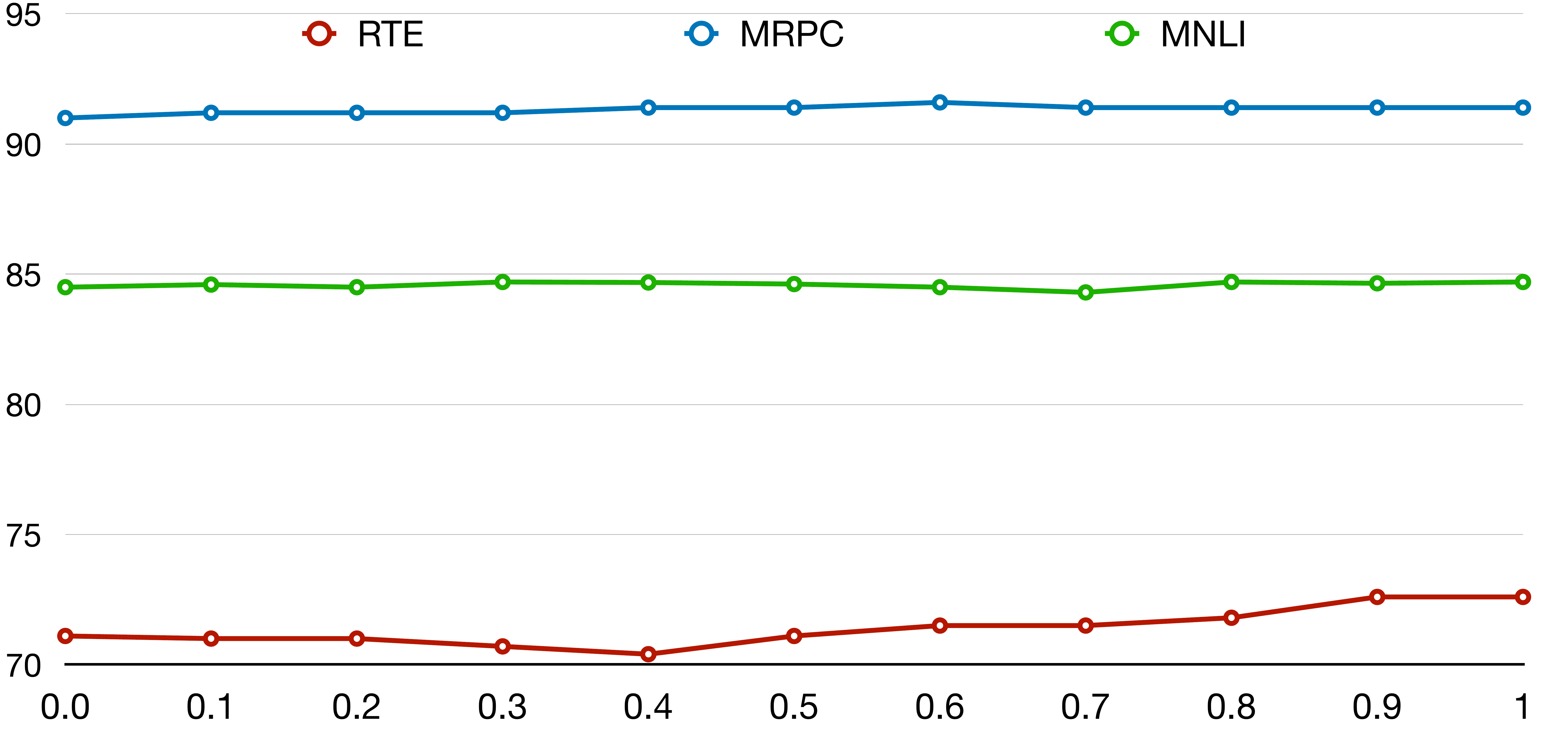}
%\caption{fig2}
\end{minipage}%
}%

\centering
\caption{KNN parameter ablation.}
\label{fig:knn-ablation}
\end{figure*}

\textbf{Importance of $K$ neighbors in KNN Classifier}

In our proposed KNN-BERT, the selection of number of nearest neighbors is a major parameter.

As seen in Fig.\ref{fig:knn-ablation}(a), the selection of $K$ is less vital when the $K$ is large enough. A very small $K$ can still have a considerable but not supreme result. 
We can summarize that a large $K$ is not necessary which can seriously save the computation cost.

\textbf{Importance of Ratio $\phi$ in Inference}

Plus, we use the combination of KNN and linear classifier during inference, therefore, the ratio $\phi$ is also important.

As seen in Fig.\ref{fig:knn-ablation}(b), the performance of the KNN classifier is better than the linear classifier while a combination of these classifiers can help gain a bigger improvement.
This indicates that KNN based classifier can achieve considerable performances while combining with the cross entropy loss optimized linear classifier is also important.

\begin{table}
\setlength{\tabcolsep}{4pt}
\centering
\small
\begin{tabular}{lcclcc}
\toprule
Methods  & RTE & MRPC  & Methods & RTE & MRPC \\
\midrule
& \multicolumn{2}{c}{\bfseries $M_m=10$ } & & \multicolumn{2}{c}{\bfseries $M_m=50$ } \\
\midrule
$M_l=1$    & 67.8 & 91.5 & $M_l=1$ & 74.7 & 91.1 \\
$M_l=10$    & 71.5 & 91.3 & $M_l=50$ & 73.8 & 91.3 \\
$M_l=20$ & 71.5 & 91.3 &   $M_l=100$ & 75.7 & 91.2 \\
\midrule
& \multicolumn{2}{c}{\bfseries $M_m=100$ } &  & \multicolumn{2}{c}{\bfseries $M_m=200$ }\\
\midrule

$M_l=1$    & 73.3 & 90.4 &$M_l=1$ & 67.1 & 90.0 \\
$M_l=100$    & 73.3 & 90.3 & $M_l=200$& 66.4 & 91.5 \\
$M_l=200$ & 67.8 & 90.2 & $M_l=400$ &  67.1 & 91.4\\

\bottomrule
\end{tabular}
\caption{Selection of $M_m$ and $M_l$ in Contrastive Learning.}
\label{tab:mn-results}
\end{table}

\textbf{Most and Least Positives Selection}

The major part of our contrastive learning framework is the selection of most and least similar positives and negatives since we aim to make use of the feature of both most and least positives to construct tighter and more distinguishable clusters.

As seen in Tab.\ref{tab:mn-results}, the selection of different $M_m$ and $M_l$ plays an important role.
Further, different tasks require different selection of $M_m$ and $M_l$.
We assume that the different size of training set requires different positives.
In this experiment, we show that using multiple positives helps achieve a considerable improvement than using a single positive sample, especially on tasks that have more diversified patterns like the RTE task.
The number of $M_m$ and $M_l$ is larger than the batch size which indicates that introducing the MoCo framework is fair and effective.
Further, we can see that different number of hard-positives $M_l$ also matters, which indicates that introducing a proper number of hard-positives is helpful in learning better representations.

\begin{figure}
\centering
\subfigure[Few-shot Linear Classifer]{
\begin{minipage}[t]{0.50\linewidth}
\includegraphics[width=1.0\linewidth]{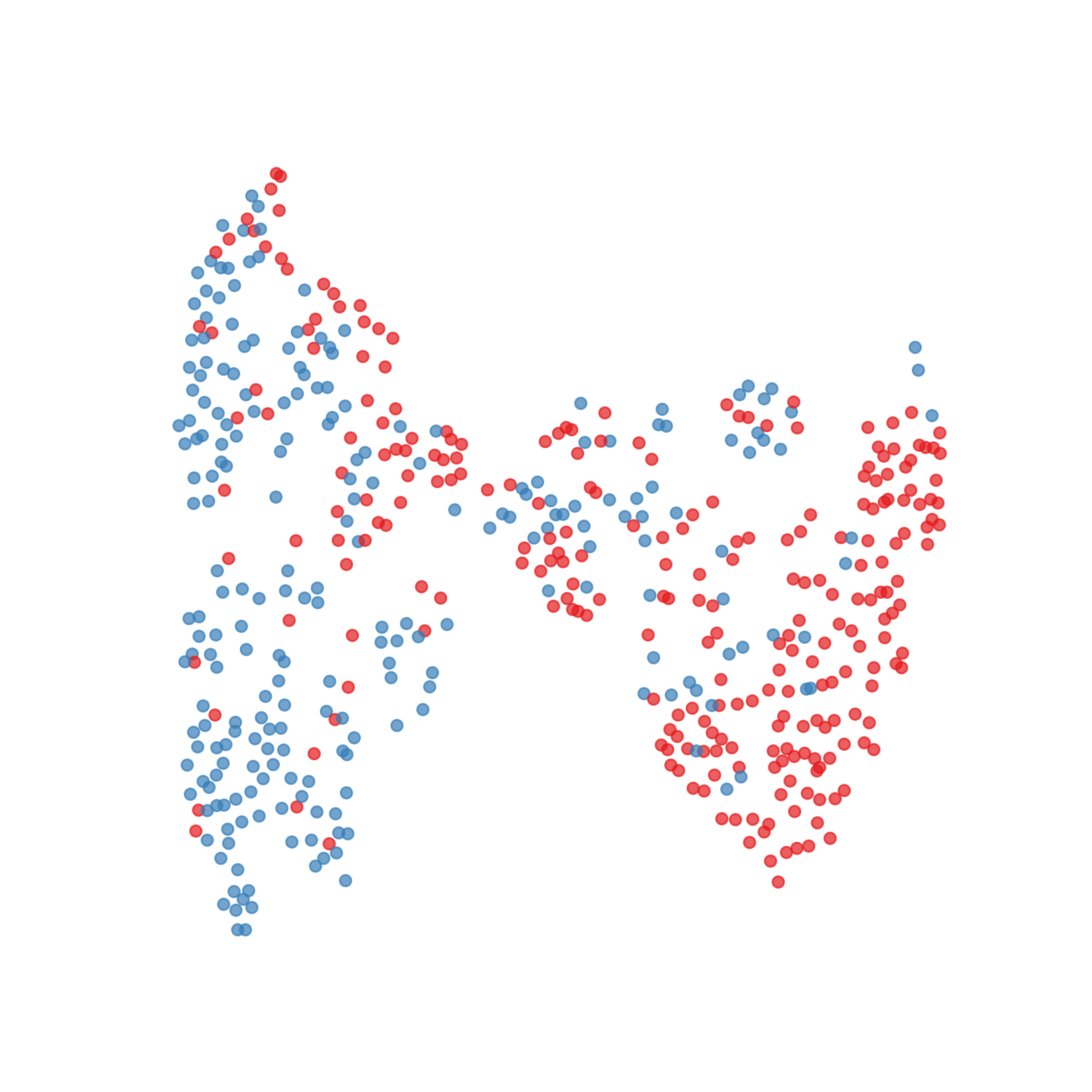}
%\caption{fig0}
\end{minipage}%
}%
\subfigure[Few-shot KNN-BERT]{
\begin{minipage}[t]{0.5\linewidth}
\includegraphics[width=1.0\linewidth]{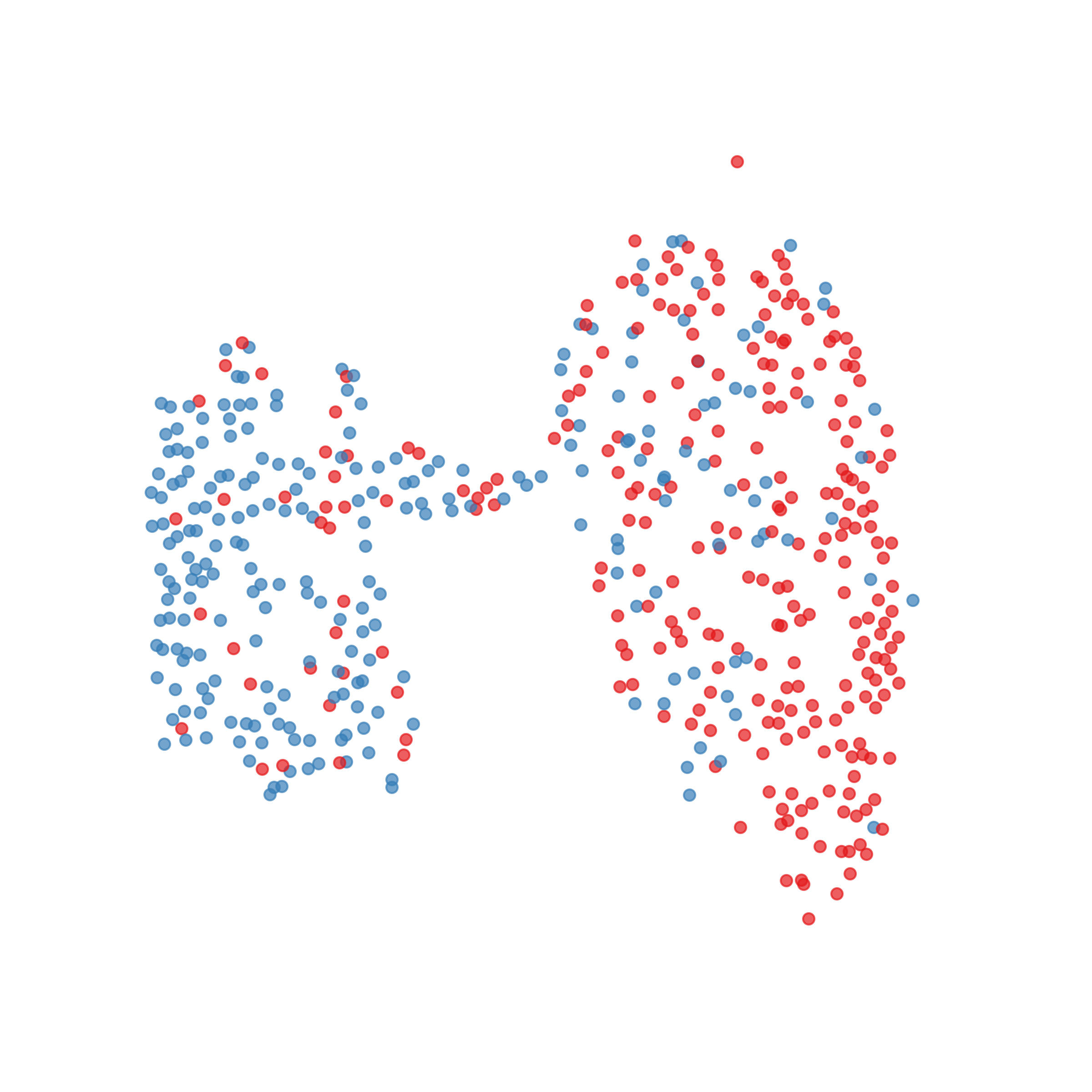}
%\caption{fig1}
\end{minipage}%
}%

\centering
\caption{TSNE Visualization, where blue dots stand for the negative samples and red dots stand for the positive samples.}
\label{fig:tsne}
\end{figure}

\subsection{Visualizations}

To explore whether the representations are actually clustered in the same class by the contrastive loss, we use the tsne visualization tool to plot the $\texttt{[CLS]}$ representations on the 100 sample fews-shot IMDB dataset using the BERT models.
We plot the representations of the original BERT fine-tuned model and our contrastive learned model using data points sampled from the development set.

As seen in Fig.\ref{fig:tsne}(a) and (b), the distance between classes is larger in the contrastive learned representations than the baseline method.
Though some of the samples are still mixed in the wrong cluster, the decision boundary is clearer and far from the other class when the representations are trained with contrastive losses.

\section{Conclusion}

In this paper, we introduce a KNN-based classifier to improve the performance of pre-trained model fine-tuning.
We utilize the traditional KNN classifier in pre-trained model fine-tuning and train clustered representations based on a supervised contrastive learning framework.
We introduce a most and least similar positive sample selection strategy based on momentum contrast framework for multiple class-wise positives and negatives in contrastive learning.
The KNN classifier can achieve higher performances on the downstream tasks while it can improve the model robustness against strong adversarial attack methods.
We are hoping that the idea of utilizing traditional the KNN classifier could provide hints for using classic machine learning methods in the state-of-the-art pre-trained language models in language understanding.

\bibliography{aaai22}

\begin{thebibliography}{50}
\providecommand{\natexlab}[1]{#1}

\bibitem[{Baevski et~al.(2020)Baevski, Zhou, Mohamed, and
  Auli}]{baevski2020wav2vec}
Baevski, A.; Zhou, H.; Mohamed, A.; and Auli, M. 2020.
\newblock wav2vec 2.0: A framework for self-supervised learning of speech
  representations.
\newblock \emph{arXiv preprint arXiv:2006.11477}.

\bibitem[{Cao et~al.(2019)Cao, Wei, Gaidon, Arechiga, and Ma}]{cao2019learning}
Cao, K.; Wei, C.; Gaidon, A.; Arechiga, N.; and Ma, T. 2019.
\newblock Learning imbalanced datasets with label-distribution-aware margin
  loss.
\newblock \emph{arXiv preprint arXiv:1906.07413}.

\bibitem[{Carlini and Wagner(2016)}]{CarliniW16a}
Carlini, N.; and Wagner, D.~A. 2016.
\newblock Towards Evaluating the Robustness of Neural Networks.
\newblock \emph{CoRR}, abs/1608.04644.

\bibitem[{Chen et~al.(2020)Chen, Kornblith, Norouzi, and
  Hinton}]{chen2020simple}
Chen, T.; Kornblith, S.; Norouzi, M.; and Hinton, G. 2020.
\newblock A simple framework for contrastive learning of visual
  representations.
\newblock In \emph{International conference on machine learning}, 1597--1607.
  PMLR.

\bibitem[{Dagan, Glickman, and Magnini(2005)}]{rte}
Dagan, I.; Glickman, O.; and Magnini, B. 2005.
\newblock The PASCAL Recognising Textual Entailment Challenge.
\newblock In \emph{Proceedings of the First International Conference on Machine
  Learning Challenges: Evaluating Predictive Uncertainty Visual Object
  Classification, and Recognizing Textual Entailment}, MLCW’05, 177–190.
  Berlin, Heidelberg: Springer-Verlag.
\newblock ISBN 3540334270.

\bibitem[{Devlin et~al.(2018)Devlin, Chang, Lee, and Toutanova}]{bert}
Devlin, J.; Chang, M.; Lee, K.; and Toutanova, K. 2018.
\newblock {BERT:} Pre-training of Deep Bidirectional Transformers for Language
  Understanding.
\newblock \emph{CoRR}, abs/1810.04805.

\bibitem[{Dodge et~al.(2020)Dodge, Ilharco, Schwartz, Farhadi, Hajishirzi, and
  Smith}]{dodge2020fine}
Dodge, J.; Ilharco, G.; Schwartz, R.; Farhadi, A.; Hajishirzi, H.; and Smith,
  N. 2020.
\newblock Fine-tuning pretrained language models: Weight initializations, data
  orders, and early stopping.
\newblock \emph{arXiv preprint arXiv:2002.06305}.

\bibitem[{Dolan and Brockett(2005)}]{dolan-brockett-2005-automatically}
Dolan, W.~B.; and Brockett, C. 2005.
\newblock Automatically Constructing a Corpus of Sentential Paraphrases.
\newblock In \emph{Proceedings of the Third International Workshop on
  Paraphrasing ({IWP}2005)}.

\bibitem[{Goodfellow, Shlens, and Szegedy(2014)}]{goodfellow2014explaining}
Goodfellow, I.~J.; Shlens, J.; and Szegedy, C. 2014.
\newblock Explaining and harnessing adversarial examples.
\newblock \emph{arXiv preprint arXiv:1412.6572}.

\bibitem[{Gunel et~al.(2020)Gunel, Du, Conneau, and
  Stoyanov}]{gunel2020supervised}
Gunel, B.; Du, J.; Conneau, A.; and Stoyanov, V. 2020.
\newblock Supervised Contrastive Learning for Pre-trained Language Model
  Fine-tuning.
\newblock \emph{arXiv preprint arXiv:2011.01403}.

\bibitem[{Gutmann and Hyv{\"a}rinen(2010)}]{gutmann2010noise}
Gutmann, M.; and Hyv{\"a}rinen, A. 2010.
\newblock Noise-contrastive estimation: A new estimation principle for
  unnormalized statistical models.
\newblock In \emph{Proceedings of the Thirteenth International Conference on
  Artificial Intelligence and Statistics}, 297--304. JMLR Workshop and
  Conference Proceedings.

\bibitem[{Hadsell, Chopra, and LeCun(2006)}]{hadsell2006dimensionality}
Hadsell, R.; Chopra, S.; and LeCun, Y. 2006.
\newblock Dimensionality reduction by learning an invariant mapping.
\newblock In \emph{2006 IEEE Computer Society Conference on Computer Vision and
  Pattern Recognition (CVPR'06)}, volume~2, 1735--1742. IEEE.

\bibitem[{He et~al.(2020)He, Fan, Wu, Xie, and Girshick}]{he2020momentum}
He, K.; Fan, H.; Wu, Y.; Xie, S.; and Girshick, R. 2020.
\newblock Momentum contrast for unsupervised visual representation learning.
\newblock In \emph{Proceedings of the IEEE/CVF Conference on Computer Vision
  and Pattern Recognition}, 9729--9738.

\bibitem[{He et~al.(2016)He, Zhang, Ren, and Sun}]{he2016deep}
He, K.; Zhang, X.; Ren, S.; and Sun, J. 2016.
\newblock Deep residual learning for image recognition.
\newblock In \emph{Proceedings of the IEEE conference on computer vision and
  pattern recognition}, 770--778.

\bibitem[{Henaff(2020)}]{henaff2020data}
Henaff, O. 2020.
\newblock Data-efficient image recognition with contrastive predictive coding.
\newblock In \emph{International Conference on Machine Learning}, 4182--4192.
  PMLR.

\bibitem[{Hjelm et~al.(2018)Hjelm, Fedorov, Lavoie-Marchildon, Grewal, Bachman,
  Trischler, and Bengio}]{hjelm2018learning}
Hjelm, R.~D.; Fedorov, A.; Lavoie-Marchildon, S.; Grewal, K.; Bachman, P.;
  Trischler, A.; and Bengio, Y. 2018.
\newblock Learning deep representations by mutual information estimation and
  maximization.
\newblock \emph{arXiv preprint arXiv:1808.06670}.

\bibitem[{Jin et~al.(2019)Jin, Jin, Zhou, and Szolovits}]{jin2019textfooler}
Jin, D.; Jin, Z.; Zhou, J.~T.; and Szolovits, P. 2019.
\newblock Is {BERT} Really Robust? Natural Language Attack on Text
  Classification and Entailment.
\newblock \emph{CoRR}, abs/1907.11932.

\bibitem[{Kassner and Sch{\"u}tze(2020)}]{kassner2020bert}
Kassner, N.; and Sch{\"u}tze, H. 2020.
\newblock BERT-kNN: Adding a kNN search component to pretrained language models
  for better QA.
\newblock \emph{arXiv preprint arXiv:2005.00766}.

\bibitem[{Khandelwal et~al.(2019)Khandelwal, Levy, Jurafsky, Zettlemoyer, and
  Lewis}]{khandelwal2019generalization}
Khandelwal, U.; Levy, O.; Jurafsky, D.; Zettlemoyer, L.; and Lewis, M. 2019.
\newblock Generalization through memorization: Nearest neighbor language
  models.
\newblock \emph{arXiv preprint arXiv:1911.00172}.

\bibitem[{Khosla et~al.(2020)Khosla, Teterwak, Wang, Sarna, Tian, Isola,
  Maschinot, Liu, and Krishnan}]{khosla2020supervised}
Khosla, P.; Teterwak, P.; Wang, C.; Sarna, A.; Tian, Y.; Isola, P.; Maschinot,
  A.; Liu, C.; and Krishnan, D. 2020.
\newblock Supervised contrastive learning.
\newblock \emph{arXiv preprint arXiv:2004.11362}.

\bibitem[{Lan et~al.(2019)Lan, Chen, Goodman, Gimpel, Sharma, and
  Soricut}]{lan2019albert}
Lan, Z.; Chen, M.; Goodman, S.; Gimpel, K.; Sharma, P.; and Soricut, R. 2019.
\newblock Albert: A lite bert for self-supervised learning of language
  representations.
\newblock \emph{arXiv preprint arXiv:1909.11942}.

\bibitem[{Li et~al.(2020)Li, Ma, Guo, Xue, and Qiu}]{li2020bert}
Li, L.; Ma, R.; Guo, Q.; Xue, X.; and Qiu, X. 2020.
\newblock Bert-attack: Adversarial attack against bert using bert.
\newblock \emph{arXiv preprint arXiv:2004.09984}.

\bibitem[{Liu et~al.(2016)Liu, Wen, Yu, and Yang}]{liu2016large}
Liu, W.; Wen, Y.; Yu, Z.; and Yang, M. 2016.
\newblock Large-margin softmax loss for convolutional neural networks.
\newblock In \emph{ICML}, volume~2, 7.

\bibitem[{Liu et~al.(2019)Liu, Ott, Goyal, Du, Joshi, Chen, Levy, Lewis,
  Zettlemoyer, and Stoyanov}]{liu2019roberta}
Liu, Y.; Ott, M.; Goyal, N.; Du, J.; Joshi, M.; Chen, D.; Levy, O.; Lewis, M.;
  Zettlemoyer, L.; and Stoyanov, V. 2019.
\newblock Roberta: A robustly optimized bert pretraining approach.
\newblock \emph{arXiv preprint arXiv:1907.11692}.

\bibitem[{Maas et~al.(2011)Maas, Daly, Pham, Huang, Ng, and
  Potts}]{maas2011learning}
Maas, A.; Daly, R.~E.; Pham, P.~T.; Huang, D.; Ng, A.~Y.; and Potts, C. 2011.
\newblock Learning word vectors for sentiment analysis.
\newblock In \emph{Proceedings of the 49th annual meeting of the association
  for computational linguistics: Human language technologies}, 142--150.

\bibitem[{Mnih and Kavukcuoglu(2013)}]{mnih2013learning}
Mnih, A.; and Kavukcuoglu, K. 2013.
\newblock Learning word embeddings efficiently with noise-contrastive
  estimation.
\newblock \emph{Advances in neural information processing systems}, 26:
  2265--2273.

\bibitem[{Morris et~al.(2020)Morris, Lifland, Yoo, Grigsby, Jin, and
  Qi}]{morris2020textattack}
Morris, J.; Lifland, E.; Yoo, J.~Y.; Grigsby, J.; Jin, D.; and Qi, Y. 2020.
\newblock TextAttack: A Framework for Adversarial Attacks, Data Augmentation,
  and Adversarial Training in NLP.
\newblock In \emph{Proceedings of the 2020 Conference on Empirical Methods in
  Natural Language Processing: System Demonstrations}, 119--126.

\bibitem[{Nar et~al.(2019)Nar, Ocal, Sastry, and Ramchandran}]{nar2019cross}
Nar, K.; Ocal, O.; Sastry, S.~S.; and Ramchandran, K. 2019.
\newblock Cross-entropy loss and low-rank features have responsibility for
  adversarial examples.
\newblock \emph{arXiv preprint arXiv:1901.08360}.

\bibitem[{Oord, Li, and Vinyals(2018)}]{oord2018representation}
Oord, A. v.~d.; Li, Y.; and Vinyals, O. 2018.
\newblock Representation learning with contrastive predictive coding.
\newblock \emph{arXiv preprint arXiv:1807.03748}.

\bibitem[{Papernot and McDaniel(2018)}]{papernot2018deep}
Papernot, N.; and McDaniel, P. 2018.
\newblock Deep k-nearest neighbors: Towards confident, interpretable and robust
  deep learning.
\newblock \emph{arXiv preprint arXiv:1803.04765}.

\bibitem[{Radford et~al.(2018)Radford, Narasimhan, Salimans, and
  Sutskever}]{radford2018improving}
Radford, A.; Narasimhan, K.; Salimans, T.; and Sutskever, I. 2018.
\newblock Improving language understanding by generative pre-training.
\newblock \emph{URL https://s3-us-west-2. amazonaws.
  com/openai-assets/researchcovers/languageunsupervised/language understanding
  paper. pdf}.

\bibitem[{Radford et~al.(2019)Radford, Wu, Child, Luan, Amodei, and
  Sutskever}]{radford2019language}
Radford, A.; Wu, J.; Child, R.; Luan, D.; Amodei, D.; and Sutskever, I. 2019.
\newblock Language Models are Unsupervised Multitask Learners.
\newblock \emph{openai}.

\bibitem[{Rajpurkar et~al.(2016)Rajpurkar, Zhang, Lopyrev, and
  Liang}]{rajpurkar-etal-2016-squad}
Rajpurkar, P.; Zhang, J.; Lopyrev, K.; and Liang, P. 2016.
\newblock {SQ}u{AD}: 100,000+ Questions for Machine Comprehension of Text.
\newblock In \emph{Proceedings of the 2016 Conference on Empirical Methods in
  Natural Language Processing}, 2383--2392. Austin, Texas: Association for
  Computational Linguistics.

\bibitem[{Rumelhart, Hinton, and Williams(1986)}]{rumelhart1986learning}
Rumelhart, D.~E.; Hinton, G.~E.; and Williams, R.~J. 1986.
\newblock Learning representations by back-propagating errors.
\newblock \emph{nature}, 323(6088): 533--536.

\bibitem[{Schroff, Kalenichenko, and Philbin(2015)}]{schroff2015facenet}
Schroff, F.; Kalenichenko, D.; and Philbin, J. 2015.
\newblock Facenet: A unified embedding for face recognition and clustering.
\newblock In \emph{Proceedings of the IEEE conference on computer vision and
  pattern recognition}, 815--823.

\bibitem[{Sermanet et~al.(2018)Sermanet, Lynch, Chebotar, Hsu, Jang, Schaal,
  Levine, and Brain}]{sermanet2018time}
Sermanet, P.; Lynch, C.; Chebotar, Y.; Hsu, J.; Jang, E.; Schaal, S.; Levine,
  S.; and Brain, G. 2018.
\newblock Time-contrastive networks: Self-supervised learning from video.
\newblock In \emph{2018 IEEE International Conference on Robotics and
  Automation (ICRA)}, 1134--1141. IEEE.

\bibitem[{Snell, Swersky, and Zemel(2017)}]{snell2017prototypical}
Snell, J.; Swersky, K.; and Zemel, R.~S. 2017.
\newblock Prototypical networks for few-shot learning.
\newblock \emph{arXiv preprint arXiv:1703.05175}.

\bibitem[{Socher et~al.(2013)Socher, Perelygin, Wu, Chuang, Manning, Ng, and
  Potts}]{socher-etal-2013-recursive}
Socher, R.; Perelygin, A.; Wu, J.; Chuang, J.; Manning, C.~D.; Ng, A.; and
  Potts, C. 2013.
\newblock Recursive Deep Models for Semantic Compositionality Over a Sentiment
  Treebank.
\newblock In \emph{Proceedings of the 2013 Conference on Empirical Methods in
  Natural Language Processing}, 1631--1642. Seattle, Washington, USA:
  Association for Computational Linguistics.

\bibitem[{Sohn(2016)}]{sohn2016improved}
Sohn, K. 2016.
\newblock Improved deep metric learning with multi-class n-pair loss objective.
\newblock In \emph{Proceedings of the 30th International Conference on Neural
  Information Processing Systems}, 1857--1865.

\bibitem[{Sukhbaatar et~al.(2014)Sukhbaatar, Bruna, Paluri, Bourdev, and
  Fergus}]{sukhbaatar2014training}
Sukhbaatar, S.; Bruna, J.; Paluri, M.; Bourdev, L.; and Fergus, R. 2014.
\newblock Training convolutional networks with noisy labels.
\newblock \emph{arXiv preprint arXiv:1406.2080}.

\bibitem[{Tschannen et~al.(2019)Tschannen, Djolonga, Rubenstein, Gelly, and
  Lucic}]{tschannen2019mutual}
Tschannen, M.; Djolonga, J.; Rubenstein, P.~K.; Gelly, S.; and Lucic, M. 2019.
\newblock On mutual information maximization for representation learning.
\newblock \emph{arXiv preprint arXiv:1907.13625}.

\bibitem[{Vaswani et~al.(2017)Vaswani, Shazeer, Parmar, Uszkoreit, Jones,
  Gomez, Kaiser, and Polosukhin}]{vaswani2017attention}
Vaswani, A.; Shazeer, N.; Parmar, N.; Uszkoreit, J.; Jones, L.; Gomez, A.~N.;
  Kaiser, {\L}.; and Polosukhin, I. 2017.
\newblock Attention is all you need.
\newblock In \emph{Advances in neural information processing systems},
  5998--6008.

\bibitem[{Wang et~al.(2018)Wang, Singh, Michael, Hill, Levy, and
  Bowman}]{wang2018glue}
Wang, A.; Singh, A.; Michael, J.; Hill, F.; Levy, O.; and Bowman, S.~R. 2018.
\newblock {GLUE}: A Multi-Task Benchmark and Analysis Platform for Natural
  Language Understanding.
\newblock ArXiv preprint 1804.07461.

\bibitem[{Weinberger and Saul(2009)}]{weinberger2009distance}
Weinberger, K.~Q.; and Saul, L.~K. 2009.
\newblock Distance metric learning for large margin nearest neighbor
  classification.
\newblock \emph{Journal of machine learning research}, 10(2).

\bibitem[{Williams, Nangia, and Bowman(2018)}]{mnli}
Williams, A.; Nangia, N.; and Bowman, S. 2018.
\newblock A Broad-Coverage Challenge Corpus for Sentence Understanding through
  Inference.
\newblock In \emph{Proceedings of the Conference of the North American Chapter
  of the Association for Computational Linguistics: Human Language
  Technologies}, 1112--1122.

\bibitem[{Wu et~al.(2018)Wu, Xiong, Yu, and Lin}]{wu2018unsupervised}
Wu, Z.; Xiong, Y.; Yu, S.~X.; and Lin, D. 2018.
\newblock Unsupervised feature learning via non-parametric instance
  discrimination.
\newblock In \emph{Proceedings of the IEEE Conference on Computer Vision and
  Pattern Recognition}, 3733--3742.

\bibitem[{Yang et~al.(2019)Yang, Dai, Yang, Carbonell, Salakhutdinov, and
  Le}]{yang2019xlnet}
Yang, Z.; Dai, Z.; Yang, Y.; Carbonell, J.; Salakhutdinov, R.; and Le, Q.~V.
  2019.
\newblock Xlnet: Generalized autoregressive pretraining for language
  understanding.
\newblock \emph{arXiv preprint arXiv:1906.08237}.

\bibitem[{Zhang et~al.(2020)Zhang, Wu, Katiyar, Weinberger, and
  Artzi}]{zhang2020revisiting}
Zhang, T.; Wu, F.; Katiyar, A.; Weinberger, K.~Q.; and Artzi, Y. 2020.
\newblock Revisiting few-sample BERT fine-tuning.
\newblock \emph{arXiv preprint arXiv:2006.05987}.

\bibitem[{Zhang, Zhao, and LeCun(2015)}]{zhang2015character}
Zhang, X.; Zhao, J.; and LeCun, Y. 2015.
\newblock Character-level convolutional networks for text classification.
\newblock In \emph{Advances in neural information processing systems},
  649--657.

\bibitem[{Zhang and Sabuncu(2018)}]{zhang2018generalized}
Zhang, Z.; and Sabuncu, M.~R. 2018.
\newblock Generalized cross entropy loss for training deep neural networks with
  noisy labels.
\newblock \emph{arXiv preprint arXiv:1805.07836}.

\end{thebibliography}

\end{document}